# Diabetic Retinopathy Classification from Retinal Images using Machine Learning Approaches

Indronil Bhattacharjee
Dept. of Computer Science and Engineering
Khulna University of Engineering & Technology (KUET)
Khulna, Bangladesh
ibprince.2489@gmail.com

Al-Mahmud
Dept. of Computer Science and Engineering
Khulna University of Engineering & Technology (KUET)
Khulna, Bangladesh
mahmud@cse.kuet.ac.bd

Tareq Mahmud
Dept. of Computer Science and Engineering
Khulna University of Engineering & Technology (KUET)
Khulna, Bangladesh
hridoytareqmahmud@gmail.com

*Abstract*— **Diabetic Retinopathy is one of the most familiar diseases and is a diabetes complication that affects eyes. Initially diabetic retinopathy may cause no symptoms or only mild vision problems. Eventually, it can cause blindness. So early detection of symptoms could help to avoid blindness. In this paper, we present some experiments on some features of Diabetic Retinopathy like properties of exudates, properties of blood vessels and properties of microaneurysm. Using the features, we can classify healthy, mild non-proliferative, moderate non-proliferative, severe non-proliferative and proliferative stage of DR. Support Vector Machine, Random Forest and Naive Bayes classifiers are used to classify the stages. Finally, Random Forest is found to be the best for higher accuracy, sensitivity and specificity of 76.5%, 77.2% and 93.3% respectively.**

**Keywords— Diabetic Retinopathy, Exudate, Blood Vessel, Microaneurysm, Random Forest.**

## I. INTRODUCTION

People suffering from diabetes can have an eye complication called diabetic retinopathy. When blood sugar levels go high, that causes harm and erosion to the blood vessels in the retina. These affected blood vessels can fatten and exude. Alternately, the vessels may have been closed, may stop flowing bloods. Sometimes unnecessary and anomalous blood vessels starts to grow on the surface of the retina. These abnormal changes can damage one's vision, sometimes may destroy fully. According to severity of the disease, DR can be classified into two main stages. (a) Non-Proliferative Diabetic Retinopathy (NPDR) and (b) Proliferative Diabetic Retinopathy (PDR).

NPDR is the initial phase of diabetic retinopathy. Many individuals with diabetes suffers from it. With NPDR, tinier blood vessels excrete and fatten the retina. When the macula expands, this has been called macular edema. This is the most familiar reason why people having diabetes leads to blindness.

In case of NPDR, the blood vessels in the retina can clogged off too. This situation is named macular ischemia. When macular ischemia happens, macula cannot get the blood supply. Intermittently some minute particles called exudates can be grown in the retina. These affects one's vision too. If anybody suffers from NPDR, his eye sight will go blurry. Furthermore, NPDR is sub classed into 3 stages, Mild, Moderate and Severe.

Proliferate DR is the most critical phase of diabetic eye disease. It occurs when the retina starts growing excessive blood vessels, which is called neovascularization. These huge numbers of vulnerable vessels often bleed into the vitreous. When they only bleed a little, a few dark floaters are found. On the other hand, when they bleed a lot, that may block the whole of the vision.

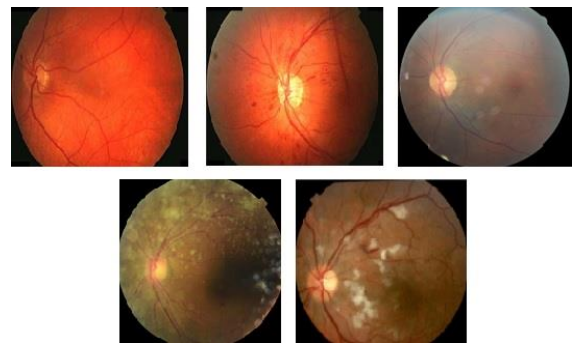

Figure I. Different stages of diabetic retinopathy (From top left): (a) Healthy Eye (b) Mild NPDR (c) Moderate NPDR (d) Severe NPDR (e) PDR

The objective of the paper are:

• Process color fundus retinal images for Diabetic Retinopathy detection.

• Extract key features from the pre-processed images.

• Detect the presence of Diabetic Retinopathy.

• Classify whether the Diabetic Retinopathy is Proliferative or Non-proliferative.

## II. THE PROPOSED SYSTEM

**Input:** Colour fundus retinal images
**Output:** Diabetic Retinopathy is not present, Mild, Moderate, Severe or PDR
**Process:**
    Step 1: Input the initial fundus image
    Step 2: Preprocess the initial image
    Step 3: Optical disk removal
    Step 4: Exudates detection
    Step 5: Blood Vessels detection
    Step 6: Microaneurysm detection

Step 7: Features extraction
Step 8: Apply to classifiers
Step 9: Classify the Diabetic Retinopathy stages
Step 10: Detect whether it is Healthy, Mild, Moderate, Severe or PDR eye

### III. METHODOLOGY

#### A. Dataset

To evaluate our method, we have used a dataset named as Diabetic Retinopathy (Resized) from Kaggle. The dataset has a total of 13402 retinal images and corresponding levels of Diabetic Retinopathy for each image.

#### B. Data Preprocessing

Data preprocessing has been done in two steps, general preprocessing for all the images and specific preprocessing for individual feature extraction.

*1) General Preprocessing:*

- *Resizing:* In this work, the sizes of the actual images in the dataset were 1024x1024 pixels. As the dataset is huge in size, the images have been stored with size 350x350 pixels for reducing the computational time.

- *Green Channel Extraction:* Preprocessing has been done with the aim of improving the contrast level of the fundus images. For contrast enhancement of the retinal images, some components like the red and blue components of the image were commonly discarded before processing. Green channel shows the clearest background contrast and greatest contrast difference between the optic disc and retinal tissue. Red channel is comparatively lighter and vascular structures are visible. The retinal vessels are lightly visible but those show less contrast than that of the green channel. Blue channel contains very little information and is comparatively noisier.

- *Contrast Limited Adaptive Histogram Equalization:* Contrast Limited Adaptive histogram equalization (CLAHE) is used for enhancing the contrast level of the images. CLAHE calculates different histograms of the image and uses these information to reallocate intensity value of image. Hence, CLAHE is sigficant for improving the regional contrast and enhancement of the edges in all the regions of an image.

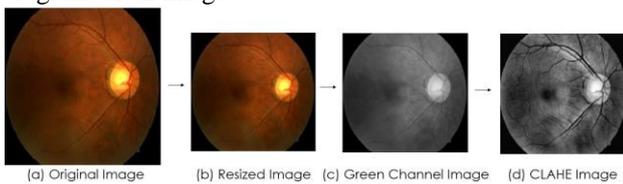

Figure II. General preprocessing

*2) Specific Preprocessing:*

- *Exudate Detection:* Firstly, optical disc has been removed using red channel of the image. Then, using 6x6 ellipse shaped structuring element, morphological dilation is applied. Non linear median filter is used for noise removal. Exudates are in high intensity values. So it has been extracted using thresholding. After applying these preprocessing, pixels having intensity value higher than 235 are set to 255 and the rest of them are set to 0 for having the clearest view. Then traversing the image, area of exudates are calculated. The images of different steps are illustrated in Figure III.

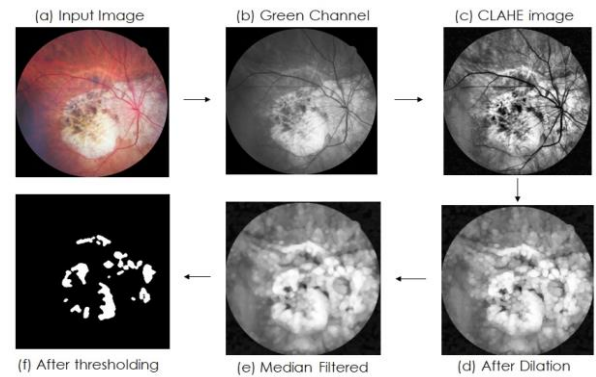

Figure III. Preprocessing for Exudate detection

- *Blood Vessel Extraction:* Blood vessel is one of the most important features for differentiating diabetic retinopathy stages. After obtaining the green channel image and improving the contrast of the image, several steps has been done for extracting blood vessel. Alternate sequential filtering (three times opening and closing) using three different sized and ellipse shaped structural element 5x5, 11x11 and 23x23 is applied on the image. Then the resultant image is subtracted from the input image. Subtracted image has lots of small noises. Those noises are removed through area parameter noise removal. Contours of each components including noises are found by finding out the contours and calculating the contour area and remove the noises which are comparatively bigger in size (200 used as reference). Then the resultant image is binarized using a threshold value. Finally the number of pixels that covers the blood vessels area are calculated. The images of different steps are illustrated in Figure IV.

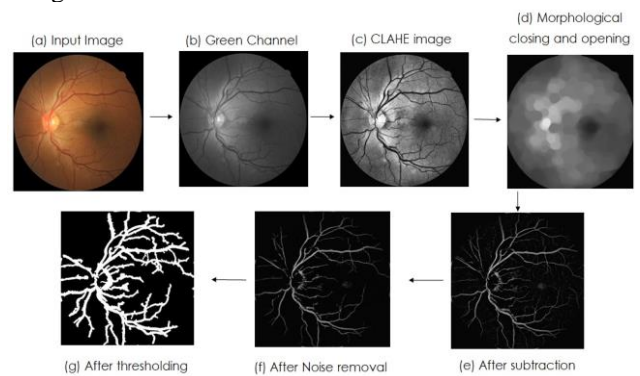

Figure IV. Preprocessing for Blood Vessel detection

- *Microaneurysm Extraction:* Green component is applied to extract microaneurysm. For better contrast, CLAHE is used. Then median filter is used for noise removal. 7x7 ellipse shaped structural element is used for morphological operation. Morphological operation erosion is applied and then the image is inverted. For joining the disjoint segments of blood vessel, morphological closing is used. Then the image has been binarized. As the blood vessel, haemorrhage and microaneurysm is having the almost same intensity, all these components will be detected altogether in the binarized image. Since microaneurysm is smaller in size, it has been extracted using contour area. The images of different steps are illustrated in Figure V.

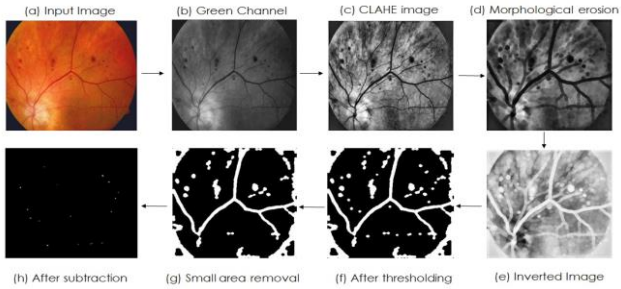

Figure V. Preprocessing for Microaneurysm detection

### C. Dataset Splitting

The dataset has been divided into two parts, where 75% as training data and 25% as test data. Therefore, 10052 training images has been used to train the model and it has been tested on 3350 images.

### D. Data Scaling

In this system, standard scalar has been used to scale all the data to limit the ranges of the variables. Using data scaling, those can be compared on the exact environments in case of all the algorithms.

### E. Selection of Features

Feature extraction will be done from preprocessed images shown in Figure III, IV and V. The features which are extracted to detect Diabetic Retinopathy are-
- Histogram of Exudates
- Zeroth Hu moment of Exudates
- Histogram of Blood Vessels
- Zeroth Hu moment of Blood Vessels
- Histogram of Microaneurysm
- Zeroth Hu moment of Microaneurysm

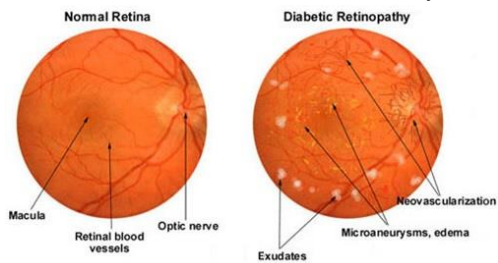

Figure VI. Differences between a normal retina and DR affected retina in terms of different features

### F. Classification

Prediction has been performed using Support Vector Machine (SVM), Random Forest (RF) and Naive Bayes classifiers.

*1) Random Forest:*

Random Forest (RF) is an ensemble tree-based learning algorithm. The RF Classifier could be a set of decision trees from an arbitrarily chosen subset of the preparing sets. It totals the votes from distinctive choice trees to choose the ultimate class of the test question. The elemental concept behind RF classifier could be a basic but capable one - the intelligence of swarms. An expansive number of moderately uncorrelated trees working as a committee will beat any of the person constituent models. Uncorrelated models can deliver gathering expectations that are more precise than any of the individual predictions. The reason for this wonderful impact is that the trees ensure each other from their personal mistakes as long as they don't always all mistakes within the same heading. Whereas a few trees may be off-base, numerous other trees will be right, so as a gather the trees are able to move within the adjusted heading.

*2) Support Vector Machine:*

SVM classify the input images into two classes such as Diabetic Retinopathy affected eye and normal eye using its features. As SVM is a binary classifier, our first task is to classify which eye is affected by Diabetic Retinopathy and which is a healthy one. After first classification, our next task is to use Support Vector Machine again. This time it is applied only on the affected ones. It will again classify which Diabetic Retinopathy is non-proliferative i.e. is in initial stage and which on is in proliferative i.e. is in severe state.

Support Vector Machine has been utilized since the SVM is based on a convex objective function that never stuck into the local maxima. The ideal hyperplane is the shape of the isolating hyperplane and the objective work of the optimization issue does not depend unequivocally on the dimensionality of the input vector but depends as it were on the inward items of two vectors. This fact permits developing the isolating hyperplanes in high-dimensional spaces.

*3) Naïve Bayes:*

Naive Bayes classifier isn't a single algorithm, but a collection of algorithms where all of them contains a common rule, that is, each match of highlights being classified is free of each other. Naive Bayes is mainly an ensemble algorithm based on Bayes' Theorem.

### G. Evaluation Metrics

Accuracy, Sensitivity and Specificity are used as evaluation metrics of the model. Accuracy, Sensitivity and Specificity are calculated using (1), (2) and (3) respectively.

$$Accuracy = \frac{True\ Positive + True\ Negative}{Total\ Number\ of\ Data} \quad (1)$$

$$Sensitivity = \frac{True\ Positive}{True\ Positive + False\ Negative} \quad (2)$$

$$Specificity = \frac{True\ Negative}{True\ Negative + False\ Positive} \qquad (3)$$

Finally, the workflow of this paper is illustrated in Figure VII

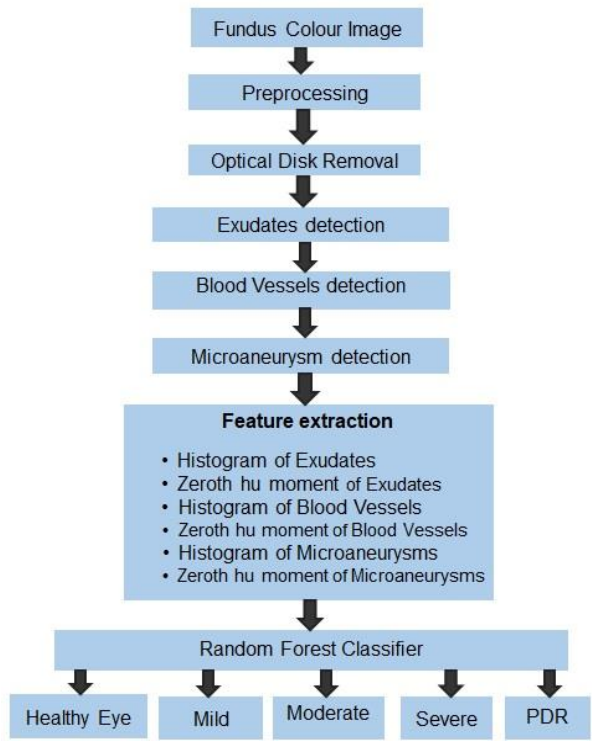

Figure VII. Flow Diagram of the system

## IV. RESULTS

Comparison of evaluation metrics for Random Forest, Support Vector Machine and Naïve Bayes classifier is shown in Table 1.

TABLE I. PERFORMANCE MEASURES OF DIFFERENT CLASSIFIERS

| Classifier | Accuracy (%) | Sensitivity (%) | Specificity (%) |
|---|---|---|---|
| Random Forest | 76.5 | 77.2 | 93.3 |
| SVM | 70.2 | 72.5 | 82.0 |
| Naïve Bayes | 67.3 | 69.4 | 75.2 |

In the experiment, Random Forest gives the highest accuracy, sensitivity and specificity. That is why Random Forest has been chosen as the best of the three classifiers used in this work.

Average sensitivity of the classification = 77.2%
Average specificity of the classification = 93.3%
Accuracy of the classification = 76.5%

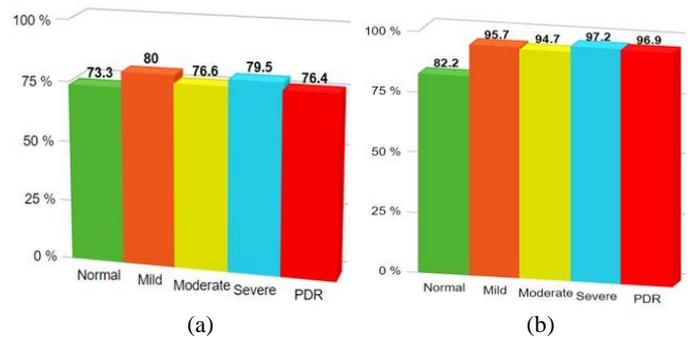

Figure VIII. Confusion Matrix for the five classes

Figure IX. Bar diagram of (a) sensitivity and (b) specificity for each class

The evaluation measures are compared with the related works in terms of number of classes, methods used, features etc.

TABLE II. COMPARISON OF DIABETIC RETINOPATHY DETECTION BY VARIOUS RESEARCHERS

| Reference | Number of Classes | Method | Accuracy (%) | Sensitivity (%) | Specificity (%) |
|---|---|---|---|---|---|
| Sinthanayothin et al. [1] | 2 | Moat operator | Not reported | 70.2 | 70.6 |
| Singalavanija et al. [2] | 2 | Exudate, Blood vessel, Microaneurysm | Not reported | 74.8 | 82.7 |
| Kahai et al. [3] | 2 | Decision support system | Not reported | 80 | 63 |
| Wang et al. [4] | 4 | Area of blood vessel | 74 | 81.7 | 92 |
| Nayak et al. [5] | 3 | Blood vessel, exudates, texture | 73.6 | 70.3 | 90 |
| Acharya et al. [6] | 5 | Higher-order spectra | 72 | 82.5 | 88.9 |
| Lim et al. [7] | 5 | Blood vessels, exudates, microaneurysm, haemorrhage | 75.9 | 80 | 86 |
| This work | 5 | Histograms and Zeroth Hu moments of blood vessel, exudates, microaneurysm | 76.5 | 77.2 | 93.3 |

Sinthanayothin et al. [1] distinguished Diabetic Retinopathy using image processing techniques from a healthy retina. In this proposed system, fundus images were preprocessed using adaptive local contrast enhancement. This method, established on a multilayer neural network, made 70.21 percent sensitivity and 70.66 percent specificity.

Kahai et al. [3] developed a system for the initial identification of the Diabetic Retinopathy. The identification system is based on a testing problem of binary-hypothesis that results only yes or no. Bayes optimization criterion was used to the raw fundus images for the initial identification of the DR. This method was able to detect the appearance of microaneurysms having sensitivity of 80 percent and specificity of 63 percent correctly.

Wang et al. [4] have classified healthy, moderate, and severe DR stages using morphological image processing approaches and a feedforward deep learning network. In this system, the existence and covering region of the components of the lesions and blood vessels are selected as the main features. The classification efficiency of this work was 74 percent, the sensitivity was 81 percent, and the specificity was 92 percent. Featuring the lesions and blood vessels, and texture parameters, they classified the input images into healthy, Moderate, and Severe DR [4].

An automatic identification system of DR was proposed by Acharya et al. [6]. They classified normal, mild, moderate, severe and PDR using the two spectral layered invariant features of higher-order spectra approaches and a Support Vector Machine classifier [8]. This work reported an accuracy of 72%, a sensitivity of 82%, and a specificity of 88%.

In this work, the fundus images are classified into five classes using the histograms and the zeroth Hu moments of the exudates, microaneurysms, and blood vessels present in the eye. Random Forest is used for the classifier. The classifier is able to identify the unknown class accurately with an efficiency of more than 76.5 percent with sensitivity 77.2 percent and specificity 93.3 percent.

Since, the instances of each class are important in this work, we have calculated the evaluation metrics by using macro-averaging method. Because macro average reveals the better scenario of the smaller classes and it is to the point and more accurate when performances on each and every classes are important equally.

## V. CONCLUSION

After studying the existing systems, we conclude that our proposed technique is successfully detecting Diabetic Retinopathy. Along with this, the proposed method is classifying into five classes of Diabetic Retinopathy. Classification has been done based on three features- area of exudates, area of blood vessel and area of microaneurysm. And using this features, we have classified into five classes as normal eye, mild NPDR, moderate NPDR, severe NPDR and PDR. Using Random Forest classifier, we have gained accuracy= 76.5%, sensitivity= 77.2% and specificity= 93.3%. The metrics we have found in this work are compared with the existing works.

In this paper, we have performed the Diabetic Retinopathy classification using Random Forest classifier with some essential features like exudates, blood vessel and microaneurysm. In future, we hope to make it work for some more classifiers like K-Nearest Neighbor classifiers and so on using some secondary features like haemorrhage also. Moreover, we can perform this classification method using larger dataset of infected eyes using neural network model in future.